\documentclass[a4paper, 10pt, conference]{ieeeconf}      
\usepackage{FG2025}
\usepackage{amsmath}
\usepackage{algorithm}
\usepackage{algpseudocode}
\usepackage{graphicx}
\usepackage{booktabs}
\usepackage{adjustbox}
\usepackage{multirow}
\usepackage{subfigure}
\usepackage{amssymb}
\usepackage{hyperref}

\FGfinalcopy 

\IEEEoverridecommandlockouts                              
\def\FGPaperID{0217} 

\title{\LARGE \bf
Towards Fair and Robust Face Parsing for Generative AI: A Multi-Objective Approach
}

\author{%
  \parbox{\textwidth}{\centering
    {\large Sophia J.~Abraham$^{1}$, Jonathan D.~Hauenstein$^{2}$, Walter J.~Scheirer$^{1}$}\\
    {\normalsize
      $^{1}$Department of Computer Science and Engineering, University of Notre Dame, Notre Dame, IN 46556\\
      $^{2}$Department of Applied and Computational Mathematics and Statistics, University of Notre Dame, Notre Dame, IN 46556
    }\\
  }
}

\begin{document}

\ifFGfinal
\thispagestyle{empty}
\pagestyle{empty}
\else
\author{Anonymous FG2025 submission\\ Paper ID \FGPaperID \\}
\pagestyle{plain}
\fi
\maketitle

\begin{abstract}
Face parsing is a fundamental task in computer vision, enabling applications such as identity verification, facial editing, and controllable image synthesis. However, existing face parsing models often lack fairness and robustness, leading to biased segmentation across demographic groups and errors under occlusions, noise, and domain shifts. These limitations affect downstream face synthesis, where segmentation biases can degrade generative model outputs.
We propose a multi-objective learning framework that optimizes accuracy, fairness, and robustness in face parsing. Our approach introduces a homotopy-based loss function that dynamically adjusts the importance of these objectives during training. To evaluate its impact, we compare multi-objective and single-objective U-Net models in a GAN-based face synthesis pipeline (Pix2PixHD). Our results show that fairness-aware and robust segmentation improves photorealism and consistency in face generation. Additionally, we conduct preliminary experiments using ControlNet, a structured conditioning model for diffusion-based synthesis, to explore how segmentation quality influences guided image generation. Our findings demonstrate that multi-objective face parsing improves demographic consistency and robustness, leading to higher-quality GAN-based synthesis.\footnote{For anonymity during the review process, source codes and trained model weights are omitted but will be made publicly available upon publication.}

\begin{figure*}[t]
    \centering
    \includegraphics[width=\textwidth]{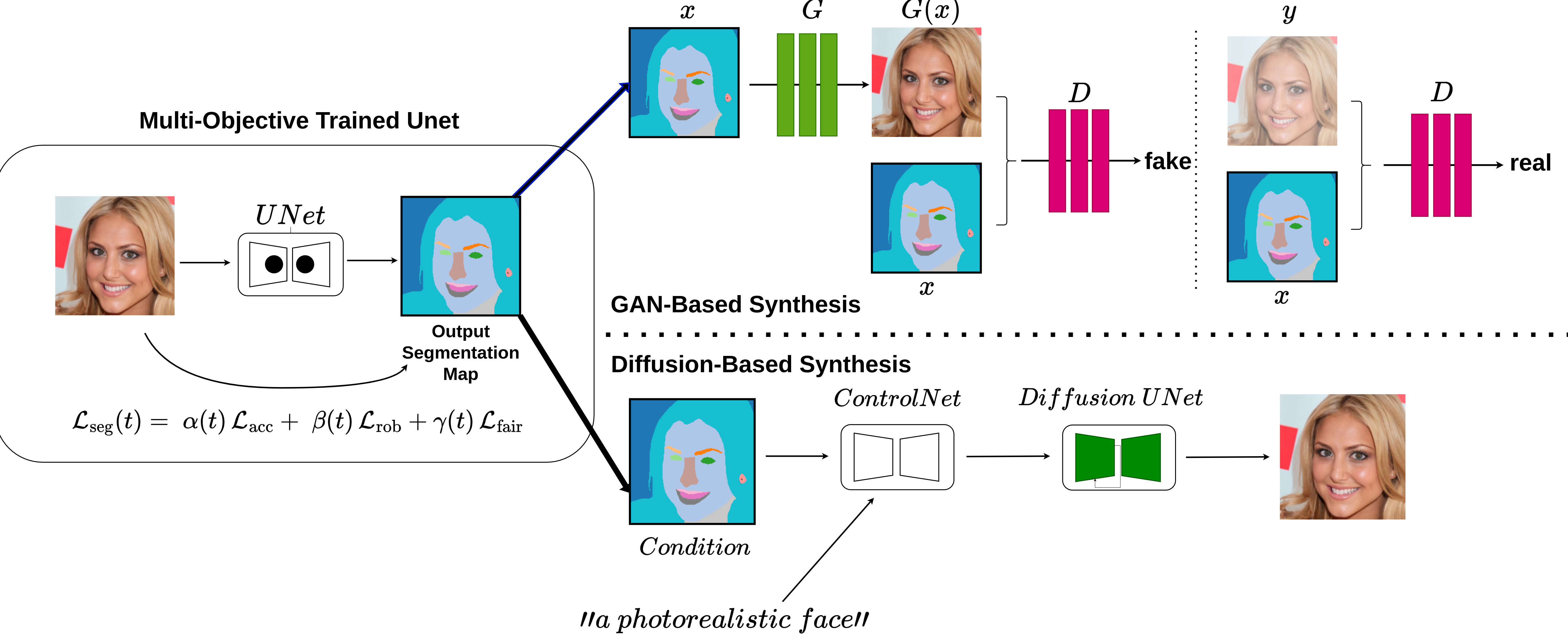}  
    \caption{
        \textbf{Overview of Our Multi-Objective Face Parsing and Synthesis Framework.} 
        Our proposed \textit{homotopy-based multi-objective learning framework} optimizes \textbf{accuracy} ($L_{\text{acc}}$), \textbf{robustness} ($L_{\text{rob}}$), and \textbf{fairness} ($L_{\text{fair}}$). 
        This framework produces \textbf{fairness-aware and robust segmentation maps}, which are used to train two generative pipelines: 
        (1) a \textbf{GAN-based synthesis model (Pix2PixHD)}, where improved segmentation enhances \textit{photorealism and demographic consistency}, and 
        (2) a \textbf{diffusion-based synthesis model (ControlNet)}, where structured parsing maps guide \textit{semantic alignment and editability}. The improved segmentation quality enhances photorealism, fairness, and robustness in generative models. Key improvements include reduced bias in GAN-generated faces and more stable semantic conditioning in diffusion synthesis.}
    \label{fig:teaser}
\end{figure*}
 
\end{abstract}

\section{Introduction}
\label{sec:intro}
Facial parsing—the segmentation of fine-grained facial components such as \textit{eyes, nose, mouth, and hair}—is a fundamental task in computer vision, supporting applications in \textit{face recognition}~\cite{wang2021deep}, \textit{augmented reality}~\cite{lee2012implementing}, and \textit{facial expression analysis}~\cite{corneanu2016survey}. Recent advances in deep learning have significantly improved segmentation accuracy~\cite{chen2017deeplab, lin2020towards}, yet existing models primarily optimize for benchmark performance while often neglecting key concerns such as: (1) \textit{fairness} across demographic groups, (2) \textit{robustness} to noise, occlusions, and domain shifts, and (3) the \textit{impact} of segmentation on downstream generative models. While face parsers may perform well on clean, well-represented data, they often degrade sharply for underrepresented demographics~\cite{buolamwini2018gender, grother2019face, park2022fair} or in challenging real-world conditions~\cite{ghiasi2014parsing, geirhos2018imagenet, hendrycks2019benchmarking}. Such biases and fragility not only reduce trust and usability in applications like \textit{identity verification} and \textit{facial editing} but also propagate into \textit{generative synthesis}, amplifying disparities in downstream tasks.

Recent efforts have explored \textit{multi-objective optimization} for general segmentation~\cite{kendall2018multi, standley2020tasks, sener2018multi} and fairness-aware approaches in facial analysis~\cite{merler2019diversity, navon2020auxiliary}. However, a unified strategy that jointly optimizes \textit{accuracy, fairness, and robustness} in face parsing remains underexplored. Furthermore, integrating fair and robust segmentation with \textit{generative models} introduces additional complexities: state-of-the-art \textbf{GAN-based}~\cite{goodfellow2014generative} and \textbf{diffusion-based} models~\cite{zhang2023adding} rely on semantically structured segmentation maps~\cite{park2019semantic, wang2018high} to generate realistic and controllable faces. If the segmentation model introduces bias or lacks robustness, these deficiencies are propagated—and often amplified—by generative models, leading to unnatural or demographically skewed outputs~\cite{menon2020pulse, tan2020improving, friedrich2023fair}. This issue is particularly pronounced in \textbf{GAN-based synthesis}, where segmentation errors cause unnatural facial reconstructions, and in \textbf{diffusion-based models} such as ControlNet, where inaccurate parsing reduces \textit{semantic alignment} and \textit{editability}.

To address these challenges, we propose a \textbf{homotopy-based multi-objective learning framework} for face parsing that explicitly balances \textit{accuracy, fairness, and robustness}. Our method dynamically adjusts training objectives over time, shifting from an accuracy-first paradigm in early training to a balanced trade-off incorporating fairness and robustness. This approach enables stronger segmentation performance across diverse demographic groups while improving resilience to \textit{occlusions, noise, and domain shifts}. Unlike prior works that optimize for fairness or robustness in isolation, our framework \textit{unifies these perspectives} within a single pipeline and systematically evaluates their impact on generative face synthesis.

To validate our approach, we integrate \textbf{multi-objective and single-objective U-Net models} into a \textbf{GAN-based face synthesis pipeline (Pix2PixHD)} and assess their impact on generative quality. We further conduct \textbf{preliminary experiments} with \textbf{ControlNet}, a structured diffusion model, to examine how segmentation quality affects \textit{guided image generation}. Our evaluations span real-world perturbations—including Gaussian noise, occlusions, blur, and lighting shifts—as well as multiple demographic groups, measuring both segmentation performance (\textit{mIoU, fairness variance}) and generative quality (\textit{Fréchet Inception Distance (FID), LPIPS similarity}~\cite{yu2021frechet, zhang2018perceptual}). Our key contributions are as follows:

\textbf{(1) Fairness-Aware Face Parsing:} We introduce a \textit{multi-objective learning framework} that explicitly optimizes \textit{accuracy, fairness, and robustness}. 

\textbf{(2) Systematic Fairness \& Robustness Evaluation:} We quantify \textit{segmentation fairness} via \textit{mIoU variance} and assess robustness under \textit{occlusions, noise, and domain shifts}. 

\textbf{(3) Impact on GAN-Based Face Synthesis:} We show that fairness-aware segmentation improves GAN-generated face quality, reducing \textit{FID scores} and enhancing perceptual realism (\textit{lower LPIPS scores}). 

\textbf{(4) Preliminary Diffusion-Based Analysis:} We explore how segmentation quality influences \textit{diffusion-based synthesis (ControlNet)}.

Our findings highlight the importance of \textbf{fair and robust face parsing} in developing \textit{bias-aware generative models} with applications in \textit{face editing, identity verification, and ethical AI deployment}. 
The remainder of this paper is structured as follows: we discuss \textit{related work} in Section~\ref{sec:relatedwork}, present our \textit{proposed method} and \textit{experiments} in Section~\ref{sec:method}, describe \textit{results} in Section~\ref{sec:results}, and conclude with \textit{implications and future research directions} in Section~\ref{sec:future}.


\section{Related Work}
\label{sec:relatedwork}
\subsection{Multi-Objective Optimization in Computer Vision}
Multi-objective optimization is widely used in computer vision to balance competing objectives such as accuracy, efficiency, and robustness~\cite{sharma2022comprehensive}. Traditional methods rely on fixed weighting schemes for loss functions, limiting adaptability across different tasks. More recent techniques, such as \textit{homotopy-based optimization}, introduce dynamic weighting mechanisms that shift priorities during training~\cite{chien2022gpu}. These methods have shown promise in solving complex optimization problems, particularly in high-dimensional polynomial systems~\cite{morgan1987computing}. However, their application to specialized areas such as \textit{face parsing and generative modeling} remains largely unexplored. Our work extends homotopy-based optimization to structured face parsing, explicitly integrating \textit{accuracy, fairness, and robustness} into a single multi-objective framework.

\subsection{Generative Adversarial Networks and Multi-Objective Training}
Generative Adversarial Networks (GANs) are widely used for tasks such as \textit{face generation, editing, and domain adaptation}~\cite{goodfellow2014generative}. Unlike diffusion models, which rely on iterative denoising, GANs synthesize high-quality images in a single forward pass, making them efficient for applications such as \textit{interactive facial editing}~\cite{karras2020analyzing}. GAN-based architectures provide structured control over facial attributes through techniques such as \textit{semantic segmentation-guided generation}~\cite{park2019semantic} and \textit{latent space manipulation}~\cite{wu2021stylespace}. 

Multi-objective training of GANs has been explored through \textit{multi-discriminator architectures} to improve stability and diversity~\cite{albuquerque2019multi} and \textit{evolutionary optimization approaches} for adversarial training~\cite{wang2019evolutionary}. However, GANs remain susceptible to \textit{mode collapse}, demographic biases, and robustness issues, particularly when trained on imbalanced datasets~\cite{tan2020improving}. Our work introduces a homotopy-based optimization framework that explicitly balances \textit{perceptual realism, semantic alignment, and demographic fairness} by leveraging segmentation maps as conditioning inputs. This allows us to systematically evaluate how fairness-aware parsing influences structured image synthesis. Additionally, we extend our analysis to \textit{diffusion-based synthesis (ControlNet)}, enabling a direct comparison between GANs and diffusion models in terms of \textit{controllability, fairness, and robustness}.

\subsection{Diffusion Models and Structured Conditioning}
Diffusion models have emerged as powerful alternatives to GANs for high-resolution image generation, achieving state-of-the-art performance in photorealistic synthesis~\cite{ho2020denoising}. Structured conditioning mechanisms, such as \textit{ControlNet}~\cite{zhang2023adding}, improve controllability by integrating external control signals, including segmentation or edge maps. While prior work has demonstrated the effectiveness of diffusion models for face synthesis~\cite{rombach2022high}, their dependence on structured inputs has not been systematically examined in the context of fairness-aware segmentation pipelines. Our study investigates the role of multi-objective face parsing in guiding diffusion-based synthesis and compares its impact against traditional GAN-based conditioning.

\subsection{Fairness in Face Parsing and Generative Models}
Fairness has been extensively studied in face recognition and classification, where demographic biases in deep learning models have been well documented~\cite{buolamwini2018gender}. However, fairness-aware segmentation remains underexplored~\cite{grother2019face}, despite evidence that segmentation models exhibit higher error rates for underrepresented demographic groups~\cite{dhar2021pass}. These disparities can propagate into downstream applications, such as attribute editing and face synthesis, amplifying biases in generative outputs.

While existing work has introduced fairness-aware regularization for generative models~\cite{tan2020improving}, few studies explicitly examine how segmentation biases affect generative synthesis pipelines. Our approach addresses this gap by incorporating fairness as an explicit training objective in face parsing and evaluating its effect on both GAN- and diffusion-based synthesis. By demonstrating how fairness-aware segmentation improves photorealism and demographic consistency, we establish a framework for more equitable face generation.

\subsection{Face Parsing for Generative Synthesis}
Face parsing, which involves segmenting facial components such as eyes, lips, and hair, plays a critical role in tasks like face editing, synthesis, and attribute manipulation~\cite{luo2012hierarchical}. Previous work has explored using segmentation maps to enhance GAN-based face editing~\cite{park2019semantic}. For instance, regional GAN inversion techniques leverage parsing maps to enable fine-grained control over facial feature editing~\cite{xu2021facecontroller}. However, existing methods prioritize accuracy without explicitly addressing fairness or robustness.

Our framework extends face parsing for generative synthesis by integrating segmentation and GAN training into a unified multi-objective pipeline. Using homotopy-based optimization, we balance \textit{realism, semantic alignment, and fairness}, ensuring that segmentation maps remain robust to variations in demographic attributes and imaging conditions.

\subsection{Robustness and Cross-Domain Generalization}
Robustness to noise, occlusion, and domain shifts remains a key challenge in vision models~\cite{geirhos2018imagenet, hendrycks2019benchmarking}. While segmentation models are often evaluated under controlled conditions, their deployment in real-world applications requires generalization across diverse datasets and imaging environments~\cite{minaee2021image}. Domain adaptation methods, such as \textit{CycleGAN}, have been explored for cross-domain transfer~\cite{zhu2017unpaired}, but they do not explicitly enforce robustness constraints in segmentation.

Our work incorporates robustness as an explicit optimization objective in face parsing, ensuring consistent performance under occlusions, noise, and cross-domain shifts. We validate our models across multiple datasets, including CelebAMask-HQ~\cite{CelebAMask-HQ}, demonstrating that multi-objective optimization enhances generalization and resilience.

\subsection{Contributions of Our Work}
While prior work has explored elements of \textit{multi-objective optimization, fairness-aware segmentation, and GAN-conditioned synthesis}, our approach is the first to integrate these into a \textit{unified homotopy-based framework}. By dynamically balancing \textit{accuracy, fairness, robustness, and semantic fidelity}, we improve face parsing performance and demonstrate its downstream impact on generative tasks.

Unlike prior methods that address fairness at the synthesis stage, our approach ensures \textit{fairness and robustness at the segmentation level}, reducing biases before they propagate into generative models. We systematically evaluate how segmentation quality affects both \textbf{GAN-based synthesis (Pix2PixHD)} and \textbf{diffusion-based synthesis (ControlNet)}, showing improvements in photorealism, demographic consistency, and structured conditioning. Our findings highlight the importance of fairness-aware segmentation for bias-aware generative modeling in face editing and synthesis.


\section{Proposed Method}
\label{sec:method}

In this section, we introduce our homotopy-based multi-objective framework for face parsing and its integration with both \textbf{GAN-based} and \textbf{diffusion-based} face editing models. We outline the problem formulation, dataset preparation, model architecture, training strategy, and evaluation pipeline, emphasizing \textbf{fairness}, \textbf{robustness}, and \textbf{semantic alignment}.

\subsection{Problem Formulation}
\label{subsec:problem_formulation}

We define the dataset \(\mathbf{X} = \{x_i\}\), where each face image is paired with a segmentation mask \( y_i \in \mathbf{Y} \), mapping to 19 facial components (e.g., hair, eyes, mouth). Demographic attributes are denoted as \(\mathbf{a}\) (e.g., \texttt{Male}, \texttt{Young}, \texttt{Wearing Hat}). Our objective is to train a segmentation function \( f_\theta(\cdot) \) that predicts \(\hat{y}_i\) while optimizing for accuracy, fairness, and robustness. Accuracy is maximized by aligning \(\hat{y}_i\) with \(y_i\) using Dice loss~\cite{sudre2017generalised}. Fairness is enforced by minimizing variance \(\mathrm{Var}(\mathrm{mIoU}_g)\) across demographic groups, ensuring equitable segmentation quality. Robustness is maintained by penalizing performance degradation (\(\mathrm{mIoU}\) drop) under input perturbations such as noise and occlusion.

\begin{algorithm}[h][t]
\caption{Multi-Objective Face Parsing (Pseudo-code)}
\label{alg:multi_objective_pseudocode}
\begin{algorithmic}[1]
\Require Homotopy function \(h(t)\) providing \((\alpha, \beta, \gamma)\) for epoch \(t\)
\For{epoch \(t = 1 \dots T\)}
    \State \((\alpha, \beta, \gamma) = h(t)\)
    \For{each batch in DataLoader}
        \State \textbf{Load} images \(\{x\}\), masks \(\{m\}\), attributes \(\{a\}\)
        \State outputs \(= f_{\theta}(x)\) \Comment{U-Net forward pass}
        \State \(\mathcal{L}_{\mathrm{acc}} = \mathrm{DiceLoss}(outputs, m)\)
        \State outputs\(_{\mathrm{noisy}} = outputs + \text{random\_noise}()\)
        \State \(\mathcal{L}_{\mathrm{rob}} = -\mathrm{mIoU}(\mathrm{softmax}(outputs_{\mathrm{noisy}}), m)\)
        \State \(\mathcal{L}_{\mathrm{fair}} = \mathrm{Var}\left[\mathrm{mIoU}_g\right]\)
        \State \(\mathcal{L}_{\text{total}} = \alpha\,\mathcal{L}_{\mathrm{acc}} + \beta\,\mathcal{L}_{\mathrm{rob}} + \gamma\,\mathcal{L}_{\mathrm{fair}}\)
        \State \textbf{Backward} and \textbf{update} \(\theta\)
    \EndFor
\EndFor
\end{algorithmic}
\end{algorithm}

\subsection{Dataset Preparation}
\label{subsec:dataset_preparation}

We employ the CelebAMask-HQ dataset \cite{CelebAMask-HQ}, divided into training, validation, and test sets. Each image and mask are resized to \(256 \times 256\) for compatibility with our U-Net architecture. Demographic attributes are extracted from annotations to compute fairness metrics.

\subsection{Model Architecture}
\label{subsec:model_architecture}

Our segmentation model utilizes a U-Net architecture with a ResNet-34 encoder pre-trained on ImageNet. It outputs 19 channels corresponding to distinct facial regions, balancing computational efficiency with high segmentation accuracy.

\subsection{Multi-Objective Training}
\label{subsec:multi_objective}

We train the U-Net segmentation models by optimizing a weighted sum of accuracy, fairness, and robustness losses, dynamically adjusted using homotopy-based scheduling. The training process is outlined in Algorithm~\ref{alg:multi_objective_pseudocode}.

\paragraph{Loss Components}
\begin{itemize}
    \item \textbf{Accuracy Loss (\(\mathcal{L}_{\mathrm{acc}}\)):} Dice loss measures the overlap between predicted and ground truth masks.
    \item \textbf{Robustness Loss (\(\mathcal{L}_{\mathrm{rob}}\)):} Negative \(\mathrm{mIoU}\) under perturbed predictions to ensure stability.
    \item \textbf{Fairness Loss (\(\mathcal{L}_{\mathrm{fair}}\)):} Variance of \(\mathrm{mIoU}\) across demographic groups to promote equitable performance.
\end{itemize}

\textbf{Alternative Fairness Computation:} We also compute per-group \(\mathrm{mIoU}\) for each demographic attribute, enabling detailed analysis of performance disparities (see Section~\ref{subsec:fairness_comparison}).

\subsection{Homotopy-Based Loss Scheduling}
\label{subsec:homotopy}

We dynamically balance the three loss components using epoch-dependent weights \(\alpha(t)\), \(\beta(t)\), and \(\gamma(t)\), ensuring \(\alpha(t) + \beta(t) + \gamma(t) = 1\). Initially, accuracy is prioritized, with weights shifting towards robustness and fairness over time. We explore three scheduling strategies:

\begin{itemize}
    \item \textbf{Linear:} \(\alpha(t)\) decreases linearly, while \(\beta(t)\) and \(\gamma(t)\) increase proportionally.
    \item \textbf{Sigmoid:} Smooth logistic transitions for gradual emphasis shifts.
    \item \textbf{Piecewise:} Abrupt changes in weight distribution at predefined training stages.
\end{itemize}

\begin{figure}[t]
    \centering
    \includegraphics[width=\columnwidth]{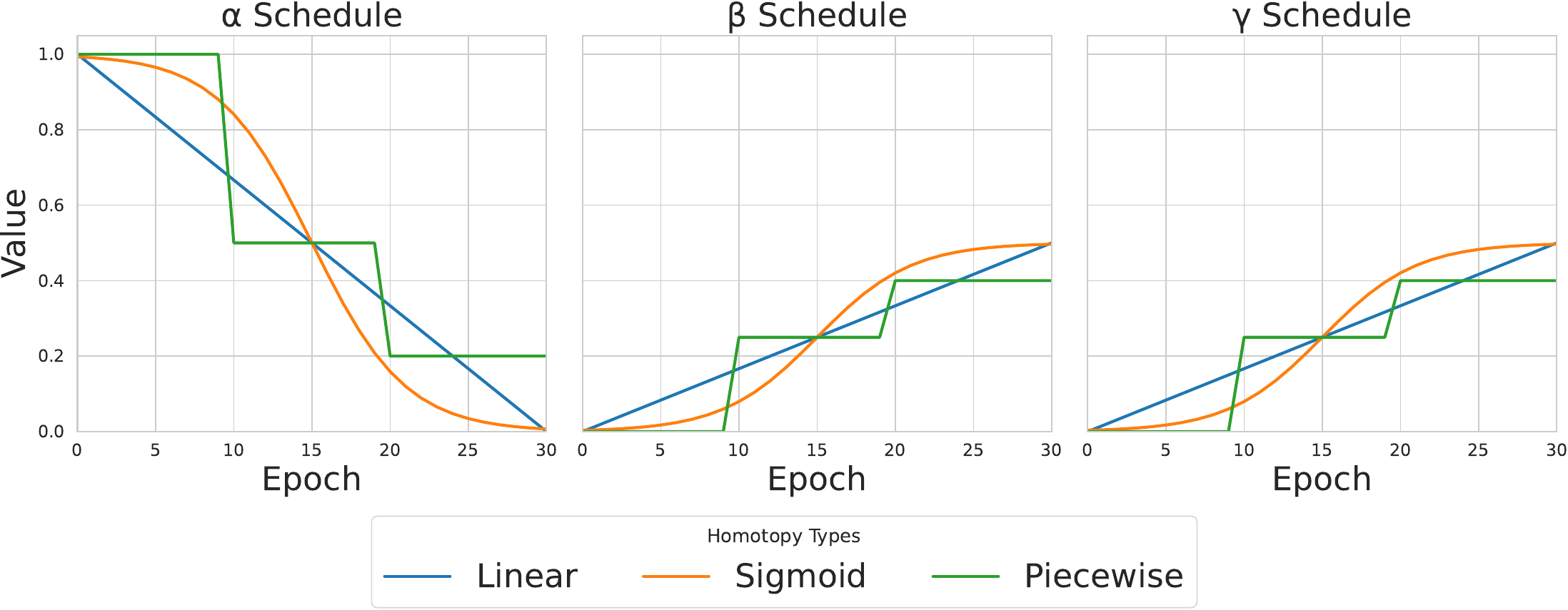}
    \caption{Comparison of \(\alpha\), \(\beta\), and \(\gamma\) schedules across three homotopy methods (Linear, Sigmoid, and Piecewise) over 30 epochs. Each subplot illustrates the evolution of a parameter (\(\alpha\), \(\beta\), or \(\gamma\)) as it adapts during training, highlighting the differences in transition dynamics across homotopy strategies. The legend below the figure identifies the homotopy method for each curve.}
    \label{fig:homotopy-schedules}
\end{figure}

Figure~\ref{fig:homotopy-schedules} illustrates the evolution of these weights across training epochs for each homotopy method.

\subsection{Integration with Generative Models}

\subsubsection{GAN-Based Face Editing}
\label{subsec:gan_integration}

We utilize the trained U-Nets to generate segmentation maps for the training and validation sets, which are then used to train a Pix2PixHD GAN. The GAN architecture comprises:

\begin{itemize}
    \item \textbf{Generator} \(G\): Transforms segmentation maps into RGB images.
    \item \textbf{Discriminator} \(D\): Distinguishes real images from generated ones.
\end{itemize}

The GAN training involves a combination of adversarial loss and pixel-level \(L_1\) reconstruction loss:
\[
\mathcal{L}_{\mathrm{GAN}} = \mathcal{L}_{\mathrm{adv}}(G, D) + \lambda \, \|\hat{x} - x\|_1,
\]
where \(\hat{x} = G(\text{segmentation\_map})\) and \(x\) is the real image.

During testing, the GAN generates images using segmentation maps from the test set produced by both single-objective and multi-objective U-Nets, enabling evaluation of how segmentation quality impacts generative performance.

\subsubsection{ControlNet-Based Face Editing}
\label{subsec:controlnet_integration}

In addition to GANs, we integrate \textbf{ControlNet} \cite{zhang2023adding} for diffusion-based face editing. ControlNet leverages segmentation maps to guide the diffusion process, enhancing image fidelity and semantic alignment. Our setup includes:

\begin{itemize}
    \item \textbf{ControlNet Model:} Pre-trained on Stable Diffusion, fine-tuned on our segmentation maps.
    \item \textbf{Diffusion Pipeline:} Combines ControlNet with a text encoder and U-Net backbone to generate photorealistic faces conditioned on segmentation maps.
\end{itemize}

\textbf{Training Procedure:} ControlNet is fine-tuned for a single epoch using segmentation maps from the training set. In diffusion-based experiments, we compare only the single-objective model with the multi-objective linear homotopy model to manage computational resources effectively. The training minimizes the standard denoising loss:
\[
\mathcal{L}_{\mathrm{ControlNet}} = \mathcal{L}_{\mathrm{denoise}},
\]
where \(\mathcal{L}_{\mathrm{denoise}}\) is the Mean Squared Error between predicted and actual noise. During testing, ControlNet generates images using test set segmentation maps from both U-Net models, allowing assessment of segmentation quality's effect on diffusion-based generation.

\subsection{Evaluation Metrics and Setup}
\label{subsec:evaluation}

\paragraph{Segmentation Metrics}  
We evaluate segmentation performance using the mean Intersection-over-Union (\(\mathrm{mIoU}\)) across 19 facial classes. Fairness is quantified by the variance \(\mathrm{Var}(\mathrm{mIoU}_g)\) across demographic groups, and robustness is assessed through performance under Gaussian noise, occlusions, and blur.

\paragraph{Generative Metrics}  
For GAN outputs, we evaluate image quality using \textbf{Fréchet Inception Distance (FID)}, which quantifies realism by comparing feature distributions between generated and real images. Additionally, \textbf{Learned Perceptual Image Patch Similarity (LPIPS)} measures perceptual similarity, where lower scores indicate greater visual resemblance to real images.

\paragraph{Implementation Details}  
All experiments are implemented in PyTorch and trained on four NVIDIA A10 GPUs using the Adam optimizer with a learning rate of \(10^{-4}\). For ControlNet, we fine-tune the pre-trained \texttt{control\_v11p\_sd15\_seg} model based on Stable Diffusion v1.5. Our pipeline supports gradient accumulation and mixed precision (FP16) for computational efficiency. Homotopy-based loss scheduling is configurable (\texttt{linear}, \texttt{sigmoid}, \texttt{piecewise}). Detailed training configurations will be released alongside our code and models to ensure reproducibility.

\paragraph{Workflow Summary}  
\begin{enumerate}
    \item \textbf{Train U-Nets:} Train single-objective and multi-objective U-Nets on the training set, validate on the validation set.
    \item \textbf{Generate Segmentation Maps:} Use trained U-Nets to produce segmentation maps for training, validation, and test sets.
    \item \textbf{Train GAN:} Train the Pix2PixHD GAN using segmentation maps from the training and validation sets.
    \item \textbf{Fine-Tune ControlNet:} Fine-tune ControlNet on training set segmentation maps for one epoch.
    \item \textbf{Generate and Evaluate Images:} Generate images using GAN and ControlNet with test set segmentation maps from both U-Net models; evaluate using FID and LPIPS.
\end{enumerate}

In the following section, we present quantitative and qualitative results demonstrating the effectiveness of our approach across various conditions and demographic groups.


\section{Results \& Discussion}
\label{sec:results}
\begin{figure}[htbp]

        \includegraphics[width=\linewidth]{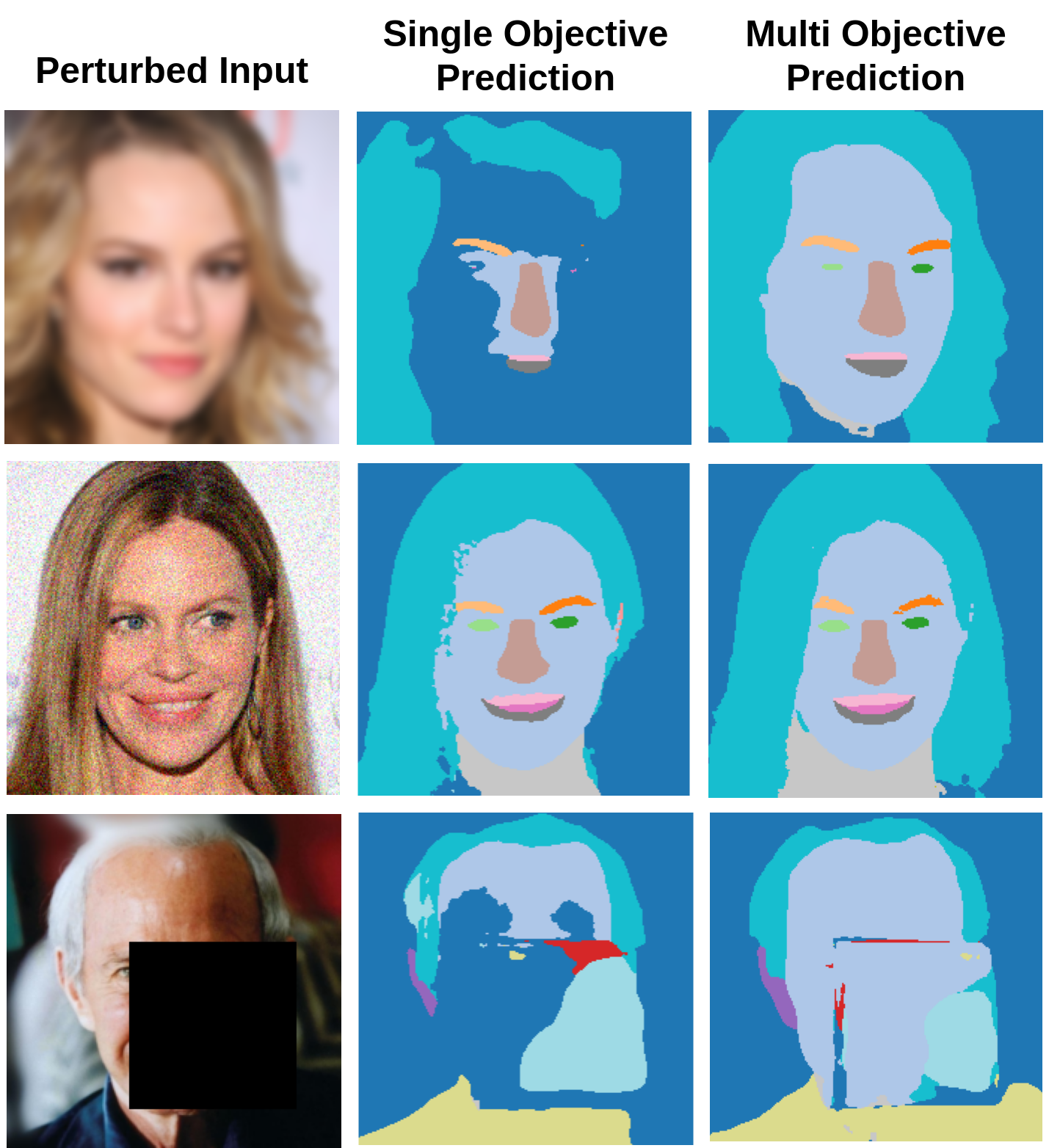}
        \label{fig:single_objective_blur}
    \caption{\textbf{Qualitative comparison of Single-Objective and Multi-Objective models under perturbations.}  
Blur (\(\text{severity} = 0.3\)), Gaussian Noise (\(\text{severity} = 0.1\)), and Occlusion (\(\text{severity} = 0.5\)) are applied to input images (first column). The Single-Objective model produces fragmented and inaccurate segmentations, especially in occluded and blurred regions. In contrast, the Multi-Objective model exhibits greater robustness, preserving facial structure despite degradations, with improved stability under occlusion.}
\label{fig:qualitative_results_comparison}
\end{figure}

\begin{figure*}[htb]
    \centering
    \includegraphics[width=\textwidth]{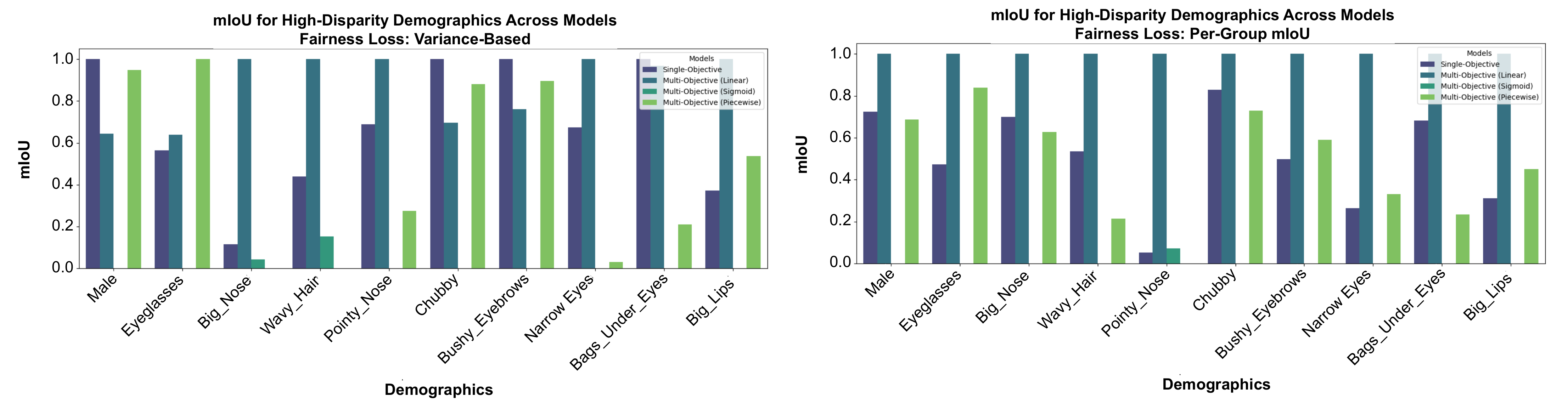}
    \caption{
        \textbf{Comparison of Fairness Loss Strategies on High-Disparity Demographics.} 
        The left plot represents the fairness variance-based approach, which minimizes the variance of per-group mIoU scores, indirectly reducing fairness gaps across demographic attributes. The right plot represents the per-group mIoU fairness loss, which explicitly tracks and optimizes fairness at a finer granularity. While the variance-based approach smooths out overall disparities, the per-group fairness loss provides better control over specific demographic attributes, ensuring higher consistency across subpopulations. Multi-objective models (Linear, Sigmoid, Piecewise) tend to provide more equitable segmentation across demographics compared to the Single-Objective baseline, though certain attributes still show variability in performance.
    }
    \label{fig:fairness_comparison}
\end{figure*}

This section presents a comprehensive evaluation of our segmentation models and their impact on both \textbf{face parsing} and \textbf{generative synthesis} (GAN and diffusion-based). We compare single-objective and multi-objective training strategies across robustness, fairness, and perceptual quality.

\subsection{Segmentation Performance}
\label{subsec:segmentation_performance}

\begin{table}[ht]
    \centering
    \caption{
        \textbf{Comparison of Segmentation Objectives on U-Net.} 
        Quantitative results comparing single-objective and multi-objective training strategies (Linear, Sigmoid, and Piecewise Homotopy) based on mean Intersection over Union (mIoU) and Dice coefficient.
    }
    \label{tab:segmentation_results_all}
    \begin{adjustbox}{max width=\textwidth}
    \begin{tabular}{@{}lcc@{}}
    \toprule
    \textbf{Objective} & \textbf{mIoU (\%)} & \textbf{Dice (\%)} \\ 
    \midrule
    Single Objective          & 73.87            & 94.46  \\
    Multi-Objective (Linear)  & \textbf{74.21}   & 94.28  \\
    Multi-Objective (Sigmoid) & 73.50            & 94.35  \\
    Multi-Objective (Piecewise) & 73.80         & 94.47  \\
    Multi-Objective (Alt. Fairness) & 73.81    & \textbf{94.49}  \\ 
    \bottomrule
    \end{tabular}
    \end{adjustbox}
\end{table}

Despite dedicating training capacity to multiple competing objectives (fairness and robustness) rather than solely optimizing for accuracy, the multi-objective models achieve segmentation performance that remains on par with or even surpasses the single-objective baseline (Table \ref{tab:segmentation_results_all}). This suggests that our homotopy-based optimization effectively balances competing goals without significantly compromising segmentation accuracy, demonstrating the feasibility of integrating fairness and robustness without sacrificing core performance.

\subsection{Robustness Analysis: Performance Under Perturbations}
\label{subsec:robustness_analysis}

To evaluate the robustness of our segmentation models, we introduce perturbations including \textbf{Gaussian noise, blur, occlusion, and salt-and-pepper noise}. We then measure mIoU degradation under increasing severity.

\begin{table*}[ht]
    \centering
    \caption{
        \textbf{Robustness of U-Net Variants Under Different Perturbations (GAN Results).}
        Each cell shows FID $\downarrow$ / LPIPS $\downarrow$. Lower FID indicates more realistic outputs, whereas LPIPS reflects perceptual distance (which can imply diversity or artifacts). 
    }
    \label{tab:robustness_results}
    \begin{adjustbox}{max width=\textwidth}
    \begin{tabular}{lccccccccl}
    \toprule
    \multirow{2}{*}{\textbf{Model}} & 
    \multicolumn{2}{c}{\textbf{Gaussian Noise}} & 
    \multicolumn{2}{c}{\textbf{Blur}} &
    \multicolumn{2}{c}{\textbf{Brightness}} &
    \multicolumn{2}{c}{\textbf{Darkness}} &
    \multirow{2}{*}{\textbf{Notes}} \\ 
    \cmidrule(r){2-9}
    & \textbf{FID $\downarrow$} & \textbf{LPIPS $\downarrow$} & \textbf{FID $\downarrow$} & \textbf{LPIPS $\downarrow$} & \textbf{FID $\downarrow$} & \textbf{LPIPS $\downarrow$} & \textbf{FID $\downarrow$} & \textbf{LPIPS $\downarrow$} & \\
    \midrule

    \textbf{U-Net (Single)} &
    363.06 & 0.435 &
    259.12 & 0.403 &
    319.57 & 0.407 &
    367.75 & 0.431 &
    Baseline segmentation \\

    \textbf{U-Net (Linear + Alt. Fairness)} &
    373.83 & 0.419 &
    211.52 & 0.390 &
    298.46 & 0.384 &
    293.71 & 0.417 &
    Linear homotopy w/ alternate fairness \\

    \textbf{U-Net (Linear)} &
    322.23 & 0.434 &
    236.44 & 0.386 &
    313.02 & 0.433 &
    285.24 & 0.425 &
    Linear homotopy approach \\

    \textbf{U-Net (Sigmoid)} &
    349.30 & 0.437 &
    208.30 & 0.412 &
    286.38 & 0.444 &
    331.36 & 0.456 &
    Sigmoid multi-objective \\

    \textbf{U-Net (Piecewise)} &
    307.16 & 0.435 &
    216.98 & 0.384 &
    330.10 & 0.439 &
    326.82 & 0.430 &
    Piecewise multi-objective \\

    \bottomrule
    \end{tabular}
    \end{adjustbox}
\end{table*}

Table~\ref{tab:robustness_results} reports the robustness of each U-Net variant to common perturbations (Gaussian noise, blur, brightness increase, darkness) within the semantic-to-image GAN framework. While the single-objective baseline exhibits high FID scores across most perturbations, it maintains moderate LPIPS values, indicating that at least part of its generated diversity may stem from artifacts or less coherent facial/gestural details. In contrast, the multi-objective piecewise and linear homotopy models generally achieve lower FID scores—particularly under noise and brightness shifts—suggesting more robust and realistic outputs. Their LPIPS values remain in a comparable range, indicating that these methods retain perceptual diversity without succumbing to as many mode distortions. Interestingly, the sigmoid variant shows strong performance against blur in terms of FID (208.30) but the highest LPIPS under darkness (0.456), possibly reflecting that it generates more varied (yet not always consistently realistic) outputs under certain perturbations. 

\begin{figure}[ht]
    \centering
    \includegraphics[width=\columnwidth]{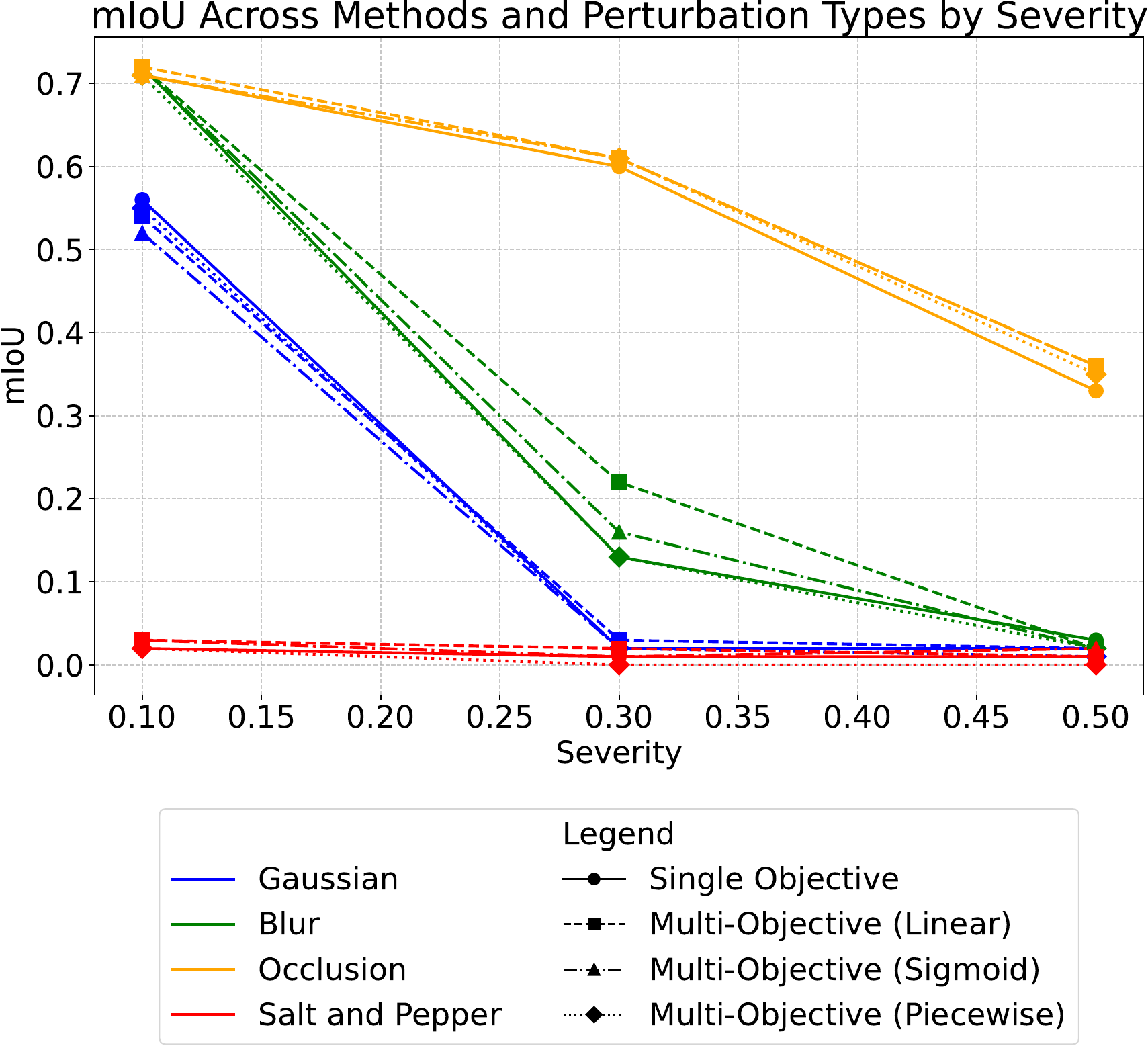}
    \caption{\textbf{Performance comparison of mIoU across methods and perturbation types under varying severities}. The plot illustrates the sensitivity of Single Objective and Multi-Objective methods (Linear, Sigmoid, and Piecewise) to perturbations, categorized by Gaussian noise, blur, occlusion, and salt-and-pepper noise. Each method is distinguished using different line styles and markers, while colors indicate the perturbation types.}
    \label{fig:miou_perturb_chart}
\end{figure}

Figure~\ref{fig:qualitative_results_comparison} illustrates segmentation performance under perturbations. The first column presents perturbed inputs, followed by predictions from the Single-Objective U-Net and the Multi-Objective U-Net (Linear). Under blur (top row), the Single-Objective model loses fine facial details, especially around the eyes and nose, leading to misalignment, whereas the Multi-Objective model maintains more cohesive structures. Gaussian noise (middle row) introduces artifacts and noisy edges in the Single-Objective model, while the Multi-Objective approach yields smoother and more stable segmentations. Occlusion (bottom row) severely disrupts Single-Objective predictions, often causing key facial regions to disappear, whereas the Multi-Objective model preserves identifiable structures, mitigating segmentation failures. These results confirm that Multi-Objective training improves segmentation resilience against real-world distortions.

In Figure \ref{fig:miou_perturb_chart} the Multi-Objective (Linear) method marginally outperforms Single Objective at mild and moderate severities. Under severe occlusion (0.5), both methods experience significant performance drops, but Multi-Objective retains a slight advantage. Salt and pepper noise sees marginal gains for Multi-Objective (Linear) at 0.1 severity (mIoU 0.03 vs. 0.02 for Single Objective), with all methods converging to very low mIoU ($\sim$0.00) at higher severities (0.3, 0.5). While Single Objective excels in specific mild perturbations, Multi-Objective methods, particularly the Linear variant, exhibit better robustness under moderate conditions, showcasing their adaptability to more challenging scenarios.

\subsection{Fairness Evaluation}
\label{subsec:fairness_eval}

\begin{table*}[ht]
    \centering
    \caption{
        \textbf{Class-wise Mean mIoU Comparison for U-Net Models.} 
        The table compares the segmentation performance of the single-objective and multi-objective trained U-Net models for each class in the CelebAMask-HQ dataset. Metrics are reported as mean Intersection-over-Union (mIoU) for 19 facial components. Higher values for each class are \textbf{bolded}.
    }
    \label{tab:unet_classwise_comparison}
    \begin{adjustbox}{max width=\textwidth}
    \begin{tabular}{@{}lccccccccccccccccccc@{}}
    \toprule
    \textbf{Model} &
    \textbf{0} &
    \textbf{1} &
    \textbf{2} &
    \textbf{3} &
    \textbf{4} &
    \textbf{5} &
    \textbf{6} &
    \textbf{7} &
    \textbf{8} &
    \textbf{9} &
    \textbf{10} &
    \textbf{11} &
    \textbf{12} &
    \textbf{13} &
    \textbf{14} &
    \textbf{15} &
    \textbf{16} &
    \textbf{17} &
    \textbf{18} \\ \midrule
    \textbf{Single Objective}      & 73.87 & 73.87 & 73.87 & 73.87 & 73.87 & 73.87 & 75.02 & 73.89 & 73.88 & 73.71 & 73.87 & 73.87 & 73.87 & 73.87 & 73.87 & 73.21 & 73.89 & 73.87 & 75.17 \\
    \textbf{Multi-Objective (Alt. Fairness)} & \textbf{74.21} & \textbf{74.21} & \textbf{74.21} & \textbf{74.21} & \textbf{74.21} & \textbf{74.21} & \textbf{75.06} & \textbf{74.23} & \textbf{74.22} & \textbf{74.09} & \textbf{74.21} & \textbf{74.18} & \textbf{74.21} & \textbf{74.21} & \textbf{74.21} & \textbf{73.49} & \textbf{74.23} & \textbf{74.21} & \textbf{75.24} \\ \bottomrule
    \end{tabular}
    \end{adjustbox}
    \vspace{1em}
\end{table*}

We measure fairness by computing performance disparities across demographic attributes. The class-wise mIoU results presented in Table~\ref{tab:unet_classwise_comparison} highlight the consistent advantages of incorporating fairness-based multi-objective training into U-Net models for face parsing. Across all 19 facial components, the multi-objective model outperforms the single-objective counterpart. Slight improvements are observed for all regions. Even for less distinctive or ambiguous classes like Class 7 (Eyebrows) and Class 15 (Accessories/Background), the multi-objective model achieves modest higher scores, with Class 15 improving from 73.21\% to 73.49\%.

\subsubsection{Comparison of Fairness Approaches}
\label{subsec:fairness_comparison}

In addition to our original fairness variance objective, we tested a refined approach that calculates per-group mIoU more explicitly (see Section~\ref{subsec:multi_objective}). Figure \ref{fig:fairness_comparison} show side-by-side comparisons of Single-Objective vs.\ Multi-Objective models across high-disparity demographic attributes. In both figures, our Multi-Objective (Linear) method generally achieves higher performance on underrepresented attributes (e.g., \texttt{Wearing\_Lipstick}, \texttt{Chubby}, \texttt{Big\_Lips}) compared to the Single-Objective baseline, though the exact mIoU values vary slightly under the new per-group logging scheme. Under this updated fairness measurement, we observe clearer demographic separations, particularly for attributes like \texttt{Eyeglasses} and \texttt{Smiling}, which highlights the subpopulations that benefit most from the Linear Homotopy weighting. Interestingly, while some categories (\texttt{Big\_Nose}, \texttt{Bags\_Under\_Eyes}) display marginally lower absolute mIoU scores, the gap between Single-Objective and Multi-Objective models becomes smaller. Finally, despite refining group-wise performance tracking, the Multi-Objective (Linear) model maintains comparable overall accuracy, indicating that this focus on fairness variance does not unduly compromise mean IoU on standard benchmarks.

\subsection{Impact on Generative Face Synthesis: GAN-Based Evaluation}
\label{subsec:gan_results}

\begin{figure}[t]
        \includegraphics[width=\linewidth]{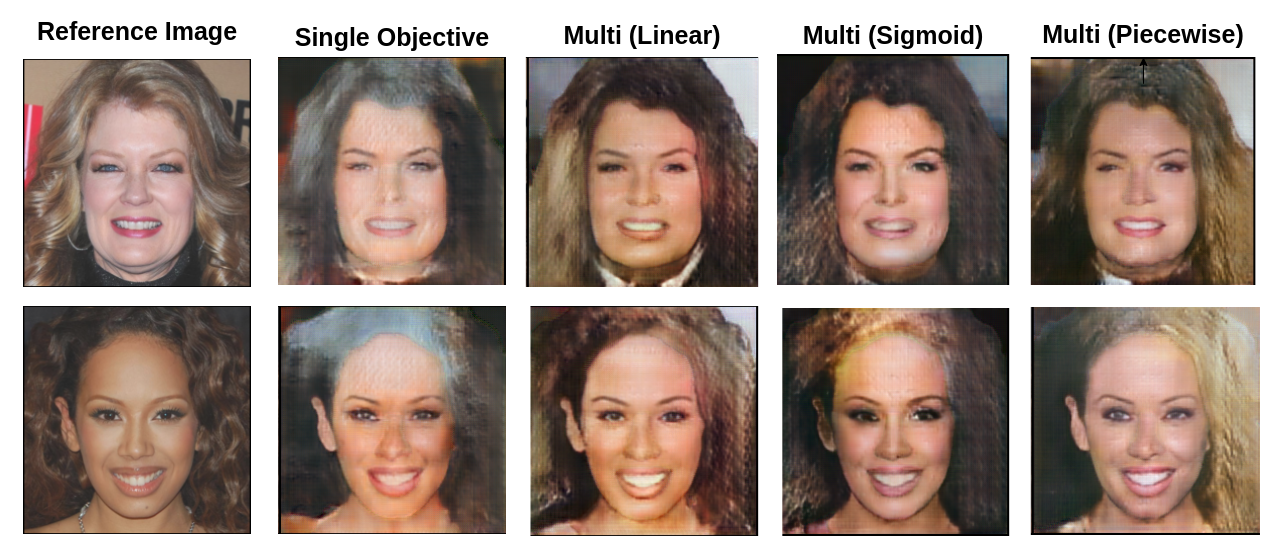}
\caption{\textbf{Impact of Segmentation Maps on GAN-Based Face Synthesis.}  
Segmentation maps from a \textbf{Single-Objective U-Net} and \textbf{Multi-Objective U-Nets} (Linear, Sigmoid, Piecewise) serve as inputs to a \textbf{Pix2Pix GAN}. Single-objective segmentation introduces inconsistencies, distorting facial details. In contrast, multi-objective segmentation improves structural coherence, yielding more natural and perceptually accurate face synthesis.}

    \label{fig:compare_composites}
\end{figure}

To analyze the downstream impact of segmentation quality, we use the generated segmentation maps as inputs to a Pix2PixHD GAN. Although we employed the same Pix2Pix-like GAN architecture for all experiments, the segmentation maps fed into the GAN were derived from U-Nets trained under distinct objectives. In the \emph{Single-Objective} case, which optimizes for raw segmentation accuracy alone, the parser occasionally misaligned facial contours—particularly around the eyes and mouth—producing vague or blurred regions in the final GAN-synthesized faces (Figure \ref{fig:qualitative_results_comparison}). By contrast, our \emph{Multi-Objective} U-Nets (Linear, Sigmoid, Piecewise) integrated fairness and robustness considerations, leading to segmentation maps with sharper boundaries and more consistent labeling of challenging facial attributes (e.g., hairlines or eyeglasses).

When these cleaner, more robust segmentation maps were passed to the same GAN, the generator more reliably reconstructed key features, yielding fewer artifacts and better overall realism. For instance, faces derived from the \emph{Multi-Objective (Sigmoid)} model exhibited reduced color bleed around the hair boundary, while the \emph{Piecewise} schedule often improved the jawline and cheek areas. Despite occasional local artifacts (e.g., minor tonal shifts), the multi-objective parses generally offered stronger geometric and semantic cues, enabling the GAN to produce final images that more faithfully mirrored the real reference. This underscores how upstream segmentation quality can impact downstream generative performance.

\begin{table}[ht]
    \centering
    \caption{\textbf{Comparison of Segmentation Models on GAN-Based and Diffusion-Based Face Synthesis.}}
    \label{tab:synthesis_results}
    \begin{adjustbox}{max width=\textwidth}
    \begin{tabular}{lcc}
    \toprule
    \multicolumn{3}{c}{\textbf{GAN-Based Face Synthesis}} \\
    \midrule
    \textbf{Segmentation Source} & \textbf{FID ↓} & \textbf{LPIPS ↓} \\ 
    \midrule
    Single-Objective U-Net       & 117.93  & 0.4419  \\
    Multi-Objective (Linear)      & 99.93   & 0.4269  \\
    Multi-Objective (Linear + Alt. Fairness)  & 107.92  & 0.4198  \\
    Multi-Objective (Sigmoid)     & 117.57  & 0.4378  \\
    Multi-Objective (Piecewise)   & \textbf{98.87}  & \textbf{0.4222}  \\
    \midrule
    \multicolumn{3}{c}{\textbf{Diffusion-Based Face Synthesis}} \\
    \midrule
    \textbf{Segmentation Source} & \textbf{FID ↓} & \textbf{LPIPS ↓} \\ 
    \midrule
    Single-Objective U-Net      & 261.01  & 0.7867  \\
    Multi-Objective U-Net (Linear)      & \textbf{257.18}  & \textbf{0.7848}  \\
    \bottomrule
    \end{tabular}
    \end{adjustbox}
\end{table}


Table~\ref{tab:synthesis_results} reports FID and LPIPS values for our U-Net-based segmentation models applied to face and gesture synthesis. While the baseline model exhibits the highest LPIPS (0.4419), a closer inspection reveals that much of this increased perceptual distance stems from undesirable artifacts and inconsistencies in facial structure, rather than meaningful diversity. This aligns with its lower overall fidelity (FID = 117.93), suggesting that higher LPIPS in this case correlates with noisy, less coherent generations rather than enhanced variation.

In contrast, the Piecewise (FID = 98.87; LPIPS = 0.4222) and Linear Homotopy (FID = 99.93; LPIPS = 0.4269) models achieve lower LPIPS while maintaining strong realism, producing more structurally accurate and perceptually consistent outputs. These models strike a crucial balance—maintaining sufficient diversity in synthesis while avoiding excessive distortions.

\subsection{Impact on Diffusion-Based Face Synthesis}
\label{subsec:diffusion_results}

To assess the role of segmentation quality in image synthesis, we integrate our segmentation maps into \textbf{ControlNet} and evaluate their impact on diffusion-based face generation. Unlike GAN-based synthesis, diffusion models rely on iterative denoising, making them particularly sensitive to structured conditioning inputs such as segmentation maps.

We conducted an initial experiment using Stable Diffusion with ControlNet, conditioned on segmentation maps from both \textbf{Single-Objective} and \textbf{Multi-Objective} U-Nets. Due to computational constraints, training was limited to \textbf{one epoch}, making this an exploratory assessment. Results in Table~\ref{tab:synthesis_results} suggest that segmentation maps from Multi-Objective models yield slight but consistent improvements in \textbf{FID} and \textbf{LPIPS}, indicating higher perceptual realism and stability. While the performance gains are modest, they demonstrate that fairness- and robustness-aware segmentation models provide more reliable conditioning for diffusion-based face generation. As this study was limited to a single training epoch, further research is needed to refine these insights. 


\begin{figure}[htbp]
    \centering
    \includegraphics[width=\linewidth]{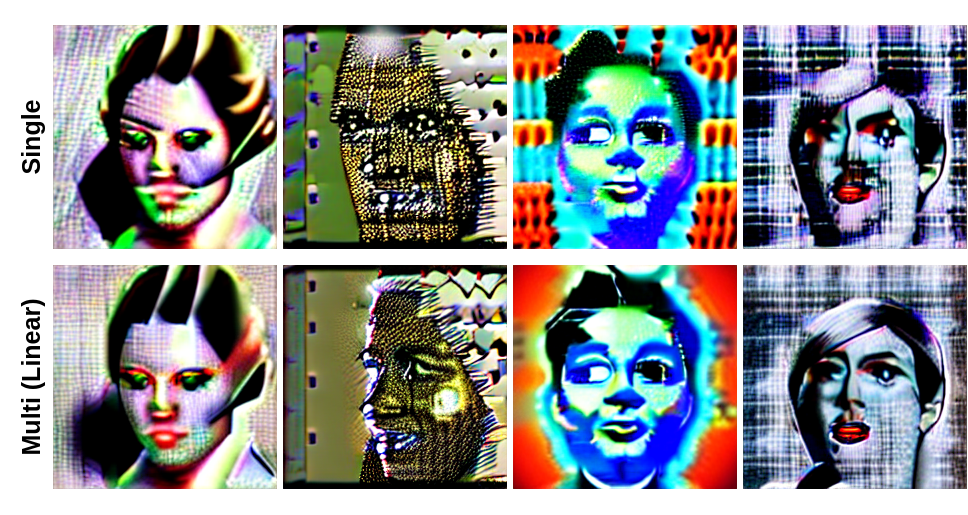}
    \caption{\textbf{Effect of Segmentation Quality on Diffusion-Based Synthesis After One Epoch of Fine-Tuning.}  
    The top row shows images generated using segmentation maps from the Single-Objective U-Net, while the bottom row corresponds to the Multi-Objective (Linear) U-Net. Despite being fine-tuned for just one epoch, the Multi-Objective model produces structurally coherent and visually consistent images, reducing artifacts and distortions in facial features. In contrast, the Single-Objective model exhibits irregular textures and geometric inconsistencies.}
    \label{fig:qualitative_results_comparison_diffusion}
\end{figure}

\vspace{-5mm}
\section{Limitations and Future Directions}
\label{sec:future}

Despite notable improvements in fairness, robustness, and segmentation quality, several challenges remain, presenting opportunities for further research.  First, the CelebAMask-HQ dataset, while diverse, remains imbalanced across demographic groups, which may limit generalization. Addressing this requires more strategic data augmentation, active reweighting, or leveraging larger, demographically-balanced datasets to further mitigate bias and enhance equitable performance. Second, our current framework treats GANs as passive consumers of segmentation maps. Incorporating \textit{bi-directional optimization}, where segmentation feedback influences GAN training, could improve both parsing fidelity and generative realism. Such an approach could be extended to diffusion models, where structured conditioning remains underexplored in fairness-aware synthesis.  

Additionally, while our method is broadly applicable beyond facial segmentation, extending it to domains such as medical imaging, autonomous perception, or video-based synthesis may require task-specific adaptations. Future research should explore domain-aware multi-objective formulations that account for context-specific biases and robustness challenges. Finally, while homotopy scheduling improves optimization efficiency, fairness-aware training introduces additional computational overhead due to subgroup evaluations. Exploring adaptive sampling strategies or efficient approximations could make large-scale deployments more feasible, especially for real-time applications.  

 Our findings underscore that multi-objective training does not impose rigid trade-offs—adaptive optimization can integrate fairness and robustness without sacrificing accuracy. By extending these ideas to broader datasets, generative frameworks, and real-world applications, future research can drive the development of more equitable and resilient vision models for AI-driven image synthesis and recognition.


\addtolength{\textheight}{-3cm}   





\section*{ETHICAL IMPACT STATEMENT}
Our research focuses on fairness-aware and robust face parsing for generative AI, addressing biases in segmentation models and their downstream impact on generative synthesis. While our work aims to mitigate demographic disparities and improve model resilience, we acknowledge potential ethical concerns related to dataset biases, misuse, and unintended societal impact.

\textbf{Potential Risks and Negative Impacts:}
Face parsing and generative models can be misused for unethical applications, such as surveillance, deepfake generation, or reinforcing demographic stereotypes. Despite our efforts to improve fairness, residual biases in datasets (e.g., CelebAMask-HQ) may persist, potentially leading to unequal model performance across demographic groups. Additionally, robustness improvements could inadvertently be leveraged to enhance adversarial facial synthesis, raising concerns about identity fraud.

\textbf{Risk-Mitigation Strategies:}
To mitigate these risks, we employ fairness-aware multi-objective training to reduce demographic disparities and systematically evaluate robustness against real-world perturbations. Our methodology prioritizes transparency and reproducibility—our dataset choices, fairness metrics, and evaluation protocols will be made publicly available to facilitate scrutiny and improvement. Furthermore, we emphasize ethical use cases, discouraging applications in deceptive or harmful generative AI practices.

\textbf{Human Subject and Data Ethics:}
Our study does not involve human subjects or personally identifiable information (PII). The datasets used (CelebAMask-HQ) are publicly available, and we adhere to all ethical guidelines concerning their use. While we acknowledge that publicly available datasets can contain biases, our methodology explicitly addresses this issue through fairness-aware training and demographic evaluation.

\textbf{Future Ethical Considerations:}
Future research should extend fairness-aware segmentation to more diverse and representative datasets, ensuring broader applicability and minimizing demographic bias. Additionally, interdisciplinary collaborations with ethicists, policymakers, and domain experts will be crucial to guiding responsible deployment and regulation of AI-generated content.

By integrating fairness and robustness into face parsing, we aim to contribute to the development of ethical, bias-aware AI models that enhance inclusivity and reliability in computer vision applications.


{\small
\bibliographystyle{ieee}
\bibliography{egbib}
}

\end{document}